# CrosswalkNet: An Optimized Deep Learning Framework for Pedestrian Crosswalk Detection in Aerial Images with High-Performance Computing

**Zubin Bhuyan, PhD**
Department of Computer Science
University of Massachusetts Lowell
1 University Avenue, Lowell, MA 01854
Email: Zubin_Bhuyan@uml.edu

**Yuanchang Xie, PhD, PE, Professor**
Department of Civil and Environmental Engineering
University of Massachusetts Lowell
1 University Avenue, Lowell, MA 01854
Email: yuanchang_xie@uml.edu

**AngkeaReach Rith**
Department of Civil and Environmental Engineering
University of Massachusetts Lowell
1 University Avenue, Lowell, MA 01854
Email: angkeareach_rith@student.uml.edu

**Xintong Yan*, PhD**
Department of Civil and Environmental Engineering
University of Massachusetts Lowell
1 University Avenue, Lowell, MA 01854
Email: xintong_yan@uml.edu

**Nasko Apostolov**
Department of Civil and Environmental Engineering
University of Massachusetts Amherst
130 Natural Resources Road, Amherst, MA 01003
Email: aapostolov@umass.edu

**Jimi Oke, PhD, Assistant Professor**
Department of Civil and Environmental Engineering
University of Massachusetts Amherst
130 Natural Resources Road, Amherst, MA 01003
Email: jboke@umass.edu

**Chengbo Ai, PhD, Associate Professor**
Department of Civil and Environmental Engineering
University of Massachusetts Amherst
130 Natural Resources Road, Amherst, MA 01003
Email: chengbo.ai@umass.edu

* Corresponding author

**ABSTRACT**
With the increasing availability of aerial and satellite imagery, deep learning presents significant potential for transportation asset management, safety analysis, and urban planning. This study introduces CrosswalkNet, a robust and efficient deep learning framework designed to detect various types of pedestrian crosswalks from 15-cm resolution aerial images. CrosswalkNet incorporates a novel detection approach that improves upon traditional object detection strategies by utilizing oriented bounding boxes (OBB), enhancing detection precision by accurately capturing crosswalks regardless of their orientation. Several optimization techniques, including Convolutional Block Attention, a dual-branch Spatial Pyramid Pooling-Fast module, and cosine annealing, are implemented to maximize performance and efficiency. A comprehensive dataset comprising over 23,000 annotated crosswalk instances is utilized to train and validate the proposed framework. The best-performing model achieves an impressive precision of 96.5% and a recall of 93.3% on aerial imagery from Massachusetts, demonstrating its accuracy and effectiveness. CrosswalkNet has also been successfully applied to datasets from New Hampshire, Virginia, and Maine without transfer learning or fine-tuning, showcasing its robustness and strong generalization capability. Additionally, the crosswalk detection results, processed using High-Performance Computing (HPC) platforms and provided in polygon shapefile format, have been shown to accelerate data processing and detection, supporting real-time analysis for safety and mobility applications. This integration offers policymakers, transportation engineers, and urban planners an effective instrument to enhance pedestrian safety and improve urban mobility.

# 1. Introduction

Crosswalks are essential components of roadway infrastructure, playing a pivotal role in pedestrian safety (Al-haideri et al., 2025; Bian et al., 2020; Pantangi et al., 2021). Accurately identifying their locations is critical for conducting pedestrian safety analysis, developing effective safety measures, and assessing pedestrian accessibility. Many states routinely collect high-resolution aerial imagery, which provides valuable information regarding locations and conditions of transportation assets. Recent advancements in deep learning, especially convolutional neural networks (CNNs), have revolutionized the analysis of large-scale image data in the field of transportation (Kortli et al., 2022; Li et al., 2024; Qingyi and Bo, 2024), enabling the automatic extraction of transportation asset information from aerial imagery. These techniques have significantly improved both the efficiency and accuracy of object detection, particularly for identifying small yet critical features like crosswalks in high-resolution aerial and satellite images (Buslaev et al., 2018; Hu et al., 2023; Sulaiman et al., 2024; Yang et al., 2019).

Several advanced object detection algorithms, such as the R-CNN (Region-based Convolutional Neural Network) suite (Gavrilescu et al., 2018; Girshick et al., 2014; He et al., 2017), the SSD (Single Shot MultiBox Detector) suite (Liu et al., 2016; Zhai et al., 2020) and the YOLO (You Only Look Once) suite (Li et al., 2022; Redmon et al., 2016; Wang et al., 2023), have been widely adopted for detecting or segmenting features from aerial images. However, the application of these models to crosswalk detection presents unique challenges due to the small size and dense distribution of crosswalks in urban environments. Additionally, extracting transportation asset data from aerial imagery involves intricate integration of deep learning with geographic information systems (GIS), as detected objects or bounding boxes must be georeferenced to hold spatial meaning. Existing deep learning models often require customization to effectively detect such small and densely packed objects (Raihan, 2023), necessitating novel approaches to improve detection accuracy and robustness.

A major challenge in processing high-resolution aerial imagery for crosswalk detection is the substantial computational power required (Berriel et al., 2017). For instance, Massachusetts is covered by over 10,000 aerial images, each containing approximately 100 million pixels. The sheer size of these images demands substantial computational resources, which can hinder the speed and scalability of processing workflows. The challenge becomes even more pronounced for larger states, such as Texas, where processing all images from a single year could take approximately 25 times longer than for Massachusetts. To address these issues, image tiling techniques, where large images are segmented into smaller, manageable pieces, are commonly employed (Chai et al., 2023; Matić et al., 2018; Unel et al., 2019). This strategy allows for faster parallel processing of image segments without compromising image quality or resolution, making it a practical solution for large-scale image analysis. However, while image tiling has proven effective for general object detection tasks, its direct application to crosswalk detection remains limited, highlighting a critical gap in existing research.

To bridge these gaps and advance the state of the art in crosswalk detection, this study introduces CrosswalkNet, a novel deep learning-based framework with oriented bounding boxes (OBBs) for accurate detection of small and densely packed crosswalks, alongside an efficient image tiling algorithm to process large datasets with scalability and computational efficiency. The proposed framework provides a robust, accurate, and scalable solution for crosswalk detection in high-resolution aerial imagery. In addition, by leveraging HPC resources, the proposed framework significantly accelerates data processing and detection, enabling near real-time analysis and decision-making. The detection results are provided in polygon shapefile format, allowing seamless integration into existing GIS-based transportation analysis workflows. The key contributions of this study are as follows:

(1) Integration of oriented bounding boxes (OBBs) for enhanced detection precision. Unlike conventional axis-aligned bounding boxes, OBBs improve detection precision by accurately capturing crosswalks regardless of their orientation, addressing the challenges posed by varied angles in urban settings.

(2) Development of an optimized image tiling algorithm for scalability. A customized image tiling approach is introduced to efficiently process large-scale datasets, facilitating high-speed parallel computation on high-performance computing (HPC) platforms while maintaining spatial integrity.

(3) Construction of a crosswalk-specific dataset. A dedicated dataset comprising over 23,000 annotated crosswalk instances is developed, incorporating domain-specific augmentation techniques to improve model generalization and address the lack of annotated data.

(4) Post-processing pipeline for crosswalk classification and refinement. A specialized post-processing module refines detection results by filtering false positives and categorizing crosswalks into meaningful types, such as "intersection," "mid-block," and "driveway," enhancing their usability for urban planning and safety applications.

The remainder of the paper is organized as follows: Section 2 provides a comprehensive review of related work on crosswalk detection and other transportation asset detection, as well as the identification of research gap and motivation. Section 3 details the data sources, dataset preparation, and annotation process. Section 4 describes the methodology for segmentation and detection, along with the architectural framework of YOLO utilized in this study. Section 5 presents a thorough discussion of the model's performance and sample results. Finally, Section 6 concludes with the significant findings, recommendations, and directions for future research.

## 2. Related work
### 2.1. Crosswalk detection

Existing crosswalk detection studies primarily rely on images captured by ground-level cameras (e.g., Haselhoff and Kummert, 2010; Kaya et al., 2023; Malbog, 2019; Murali and Coughlan, 2013; Tian et al., 2021; Tümen and Ergen, 2020; Zhai et al., 2015; Zhang et al., 2022). These studies benefit from the availability of high-resolution images with detailed perspectives of road markings, making them particularly effective for identifying crosswalks in specific localized areas. However, such methods are inherently constrained by their limited spatial coverage and dependence on camera placements (Kaufmann et al., 2018; Xing et al., 2019). Additionally, ground-level cameras primarily provide information from a micro-level perspective, focusing on localized details but lacking the macroscopic insight required for city- or region-wide urban planning and transportation analysis. Consequently, these studies are less suited for large-scale GIS-based analyses, where integration with spatial data is crucial.

Aerial and satellite imagery provides a promising alternative by enabling crosswalk detection at broader spatial scales. Several studies have explored the potential of aerial imagery for this purpose (Antwi and Takyi, 2024; Berriel et al., 2017; Luttrell et al., 2024; Verma and Ukkusuri, 2024; Xie et al., 2023; Yi et al., 2021). For instance, Berriel et al. (2017) pioneered the use of deep learning for crosswalk detection in satellite images, developing a dataset covering multiple cities and continents to train a VGG neural network. However, their approach primarily focused on binary classification (i.e., determining the presence of crosswalks within an image patch), which limits its practical utility for GIS-based safety analysis.

More recent studies have adopted object detection techniques such as YOLO (Verma and Ukkusuri, 2024; Luttrell et al., 2024) to detect crosswalks in aerial imagery. While these models achieved promising accuracy, they relied on axis-aligned bounding boxes (AABs), which struggle to accurately capture crosswalks with varying orientations in complex urban environments. Moreover, most of these approaches utilized small image sizes (e.g., 256×256 or 640×640 pixels), which are suboptimal for high-resolution aerial imagery and necessitate extensive preprocessing to compensate for scale limitations.

Xie et al. (2023) introduced a system that detects intersection markings and evaluates degradation using aerial imagery, employing Box Boundary-Aware Vectors (BBAVectors) (Yi et al., 2021) for oriented detection. Despite their progress, their focus was on degradation assessment of pavement markings rather than precise crosswalk detection, leaving a gap in the development of frameworks tailored for large-scale transportation planning and safety applications.

### 2.2. Other transportation asset detection

In addition to crosswalks, extensive research has been conducted on detecting other transportation assets using deep learning applied to aerial and ground-level imagery. Some studies leverage aerial and satellite images (e.g., Bastani et al., 2018; Brkić et al., 2023; Hosseini et al., 2023; Kumar et al., 2019; Li and Goldberg, 2018; Mahaarachchi et al., 2023), while others utilize street-level images from platforms like

Google Street View (e.g., Balali et al., 2015; Campbell et al., 2019; Kong et al., 2022; Ren et al., 2023; Han, 2025).

Using satellite imagery, Brkić et al. (2023) developed a model to identify various transportation features, including refugee islands, school zones, and multiple classes of crosswalks. Although their model achieved high precision (95.7%–98.8%), it exhibited lower recall values (75.9%–90.3%), highlighting challenges in comprehensive detection across diverse environments. Hosseini et al., (2023) introduced TILE2NET, an end-to-end semantic segmentation tool to extract sidewalks, crosswalks, and footpath polygons from aerial imagery, demonstrating potential for large-scale urban planning applications. Mahaarachchi et al. (2023) applied a mask-RCNN framework to detect parking spaces in aerial images, achieving an accuracy of 97.7%. However, scalability issues persisted due to the computational complexity of processing large aerial datasets.

Regarding ground-level images, Campbell et al. (2019) employed the SSD MobileNet framework to detect traffic signs from Google Street View images with a detection accuracy of 95.63%. Ren et al. (2023) proposed an improved YOLOv5-based pavement damage detection approach using street-view images, achieving an average precision of 79.8%. Han (2025) employed PSPNet for road assets in street-view images including sidewalks, roads, traffic signs and so on. By pretraining the model on the ADE20k dataset, a notable 80.8% accuracy in pixel-wise predictions is obtained. However, as discussed earlier, ground-level imagery is inherently limited by its localized coverage and inability to capture the broader transportation network, making it unsuitable for macro-level planning and analysis.

Several studies have attempted to combine aerial and street-view imagery for a more comprehensive transportation asset assessment (e.g., Guo et al., 2024; Hoffmann et al., 2019; Ning et al., 2022). For instance, Ning et al. (2022) employed YOLACT (You Only Look At CoefficienTs) to extract sidewalks from aerial images, and then use pre-trained models to restore occluded and missing sidewalks from street view images. Guo et al. (2024) introduced the Two-Branch Contextual Feature-Guided Converged Network (TCFGC-Net) to conduct segementation of different elements such as road, sidewalk and building from satellite imagery and dynamic continuous features from street view imagery. However, such hybrid approaches introduce challenges in data integration and alignment across different sources.

## 2.3. Research gap and motivation

Despite significant progress in aerial imagery-based transportation asset detection, several critical challenges remain. One major limitation of existing methods is their reliance on axis-aligned bounding boxes (AABs), which are not well-suited for accurately detecting crosswalks with varying orientations, especially in complex urban environments. This constraint often leads to imprecise localization and misclassification of crosswalks. To overcome this issue, our study incorporates oriented bounding boxes (OBBs), which provide a more accurate representation of crosswalks, regardless of their orientation.

Another key challenge lies in the scalability of detection methods. Many prior studies process aerial imagery in small image patches, which, although computationally manageable, significantly limit the efficiency of analyzing high-resolution datasets. To address this, our framework introduces an optimized image tiling algorithm, which facilitates seamless parallel processing using high-performance computing (HPC) platforms. This approach ensures efficient data handling without compromising spatial integrity, making it suitable for large-scale applications.

In addition to scalability concerns, most existing research focuses solely on determining the presence or absence of crosswalks, offering limited practical value for urban planning and transportation management. Our framework aims to fill this gap by not only detecting crosswalks but also classifying them into functional categories such as "intersection," "mid-block," and "driveway." This categorization provides deeper insights that can support safety assessments and infrastructure planning efforts.

Finally, ensuring generalization across diverse geographic areas remains a critical challenge for crosswalk detection models. Many existing methods are trained and validated within a single geographic region, resulting in limited adaptability to different environments. In contrast, our framework is rigorously tested across datasets from Massachusetts, New Hampshire, Virginia, and Maine. This evaluation demonstrates the model's robustness and ability to generalize effectively across varying urban settings without requiring additional fine-tuning or transfer learning.

## 3. Data
### 3.1. Data Source
The aerial imagery used in this study was acquired from the Massachusetts Bureau of Geographic Information (MassGIS). Images for 2019 and 2021 were downloaded from the MassGIS website. The image dataset for each year comprises over 10,000 high-resolution images (tiles). Each image contains 100 million pixels (10,000 x 10,000 pixels), with each pixel representing about 6 inches (15 centimeters) on the ground. This resolution provides sufficient detail for identifying small-sized features such as pedestrian crosswalks.

### 3.2. Dataset preparation
Due to the large size of the original images, a technique called "slicing" was employed, in which the images were divided into smaller, overlapping segments or "patches" measuring 1024x1024 pixels. The full image was cut into manageable squares that could be processed individually by the deep learning model. The overlapping patches were designed to share some area with adjacent segments. This approach was found to be useful in managing the computational load more effectively, ensuring that the deep learning model could process the data without encountering memory constraints. The overlapping regions were particularly beneficial as they helped capture contextual information around the edges of each patch and improved the model's ability to detect and analyze features spanning multiple patches. Experiments were conducted with other patch sizes, and 1024 pixels were determined to provide the best performance in terms of accuracy and processing speed.

Furthermore, the original three-channel RGB images were converted into grayscale ones to simplify them to a single-band input for deep learning models, potentially reducing the computational load. While this conversion was found to decrease memory requirements and accelerate processing speed for faster training and inference times, it also involved a trade-off with the loss of color information, which can be critical for accurately detecting pedestrian crosswalks. Therefore, both the original RGB and grayscale images were considered in this research.

### 3.3. Data annotation
Annotation was performed by labeling specific features or objects within images to provide ground truth data for training deep learning models. The patch images of 1024x1024 pixels were annotated with detailed labels to facilitate precise training of deep learning models for crosswalk detection. The annotated dataset encompassed 44 tiles of aerial imagery, labeled for three distinct types of crosswalks: zebra/continental/ladder crosswalks, crosswalks marked by two parallel solid lines, and crosswalks with solid pavement markings. A few examples of different types of crosswalks are shown in **Fig. 1**.

Zebra, continental, and ladder crosswalks, characterized by their broad, white stripes, were categorized into the same group due to their similarity and common usage. Crosswalks with two parallel solid lines were also widely adopted. Sometimes, the area between the two lines was painted in solid colors. Since they all had two parallel lines, both cases were categorized into the same annotation group. Crosswalks with solid pavement markings, which were sometimes physically elevated to slow down traffic and enhance pedestrian safety, were considered. They were less common and typically used in residential areas. Given the limited samples of such crosswalks and the primary focus on major roads, they were later excluded from the analysis.

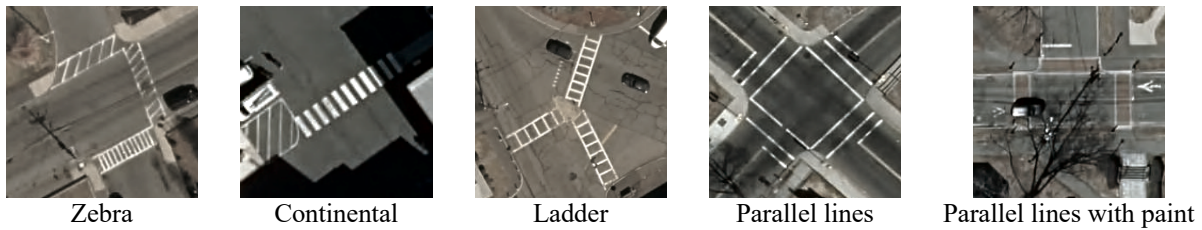

| Zebra | Continental | Ladder | Parallel lines | Parallel lines with paint |

**Fig. 1.** Examples of various types of crosswalks from our dataset.

In total, we annotated 23,938 crosswalks. Of these, 18,885 crosswalks were utilized for model training, while 5,053 were set aside for validation purposes. We also augmented the training dataset using some standard data augmentation techniques.

## 4. Methodology
### 4.1. Segmentation v.s. detection

In the development of CrosswalkNet, both segmentation and detection approaches were evaluated for pedestrian crosswalk detection. Several models, including Mask R-CNN, Faster R-CNN, U-Net, DeepLabv3+, and YOLOv5—commonly available in commercial GIS tools—were tested to determine the most suitable approach. Detection models were observed to significantly outperform segmentation models in terms of computational efficiency during both training and inference. Given the need to process over 10,000 aerial images annually for Massachusetts, computational efficiency was identified as a critical factor. However, it was found that traditional detection models, which generate horizontal bounding boxes (HBBs), often include non-crosswalk areas and produce overlapping detections, complicating the classification of different crosswalk types when integrated with GIS-based road network data

Conversely, segmentation models such as DeepLabv3+ provided pixel-level classification, offering higher localization accuracy. However, their slower processing speeds and extensive post-processing requirements were identified as major drawbacks. The process of converting classified pixels into polygons required additional efforts and often led to the merging of adjacent pedestrian crosswalks into a single large polygon, making it challenging to obtain accurate crosswalk counts for subsequent analyses.

To overcome these challenges, the CrosswalkNet detection framework was developed based on YOLOv8. The model leverages oriented bounding boxes (OBBs) to achieve precise crosswalk localization, effectively addressing the limitations of HBB-based methods by reducing false positive regions and enhancing spatial accuracy. Furthermore, seamless integration with GIS platforms was achieved, allowing detected crosswalks to be exported in polygon shapefile format for further spatial analysis.

CrosswalkNet was developed and implemented using Python, with training, testing, and inference conducted on NVIDIA RTX 4090 and A100 GPUs, ensuring the efficient processing of large-scale datasets. Additionally, the training process was scaled up using the Unity HPC cluster at the Massachusetts Green High Performance Computing Center (MGHPCC). Unity comprises over 20,000 CPU cores and 1,500 NVIDIA GPUs (including A100 and V100 models), operating on Ubuntu 20.04 LTS with Slurm job scheduling to facilitate optimal resource allocation and parallel processing of aerial imagery. This setup ensures the efficient execution of large-scale crosswalk detection tasks.

### 4.2. CrosswalkNet: an enhanced deep learning framework for crosswalk detection

CrosswalkNet is an advanced deep learning framework specifically designed to address the unique challenges of pedestrian crosswalk detection in aerial imagery. As discussed in Section 4.1, extensive experimentation was conducted with various object detection and segmentation architectures. The YOLO (You Only Look Once) framework was ultimately selected as the most suitable due to its efficiency in real-time applications and high detection accuracy.

A comprehensive evaluation revealed that YOLOv8 provides the optimal balance between precision and inference speed, making it highly effective for processing large-scale aerial datasets. While subsequent versions, such as YOLOv9, introduced architectural modifications, they did not offer significant performance improvements for this task. Moreover, the latest release, YOLOv10, was not utilized in this study due to the lack of official support for oriented bounding boxes (OBBs), which are essential for accurately detecting crosswalks at various orientations.

The architecture of CrosswalkNet is designed to address the unique challenges of pedestrian crosswalk detection in aerial imagery. It consists of several key components (shown in **Fig.2**): an input layer for data preprocessing, a customized backbone network for hierarchical feature extraction, a refined neck network to aggregate multi-scale features, and a detection head responsible for generating accurate crosswalk predictions. Post-processing steps, such as non-maximum suppression (NMS), are applied to refine detection outputs, improving precision and reducing false positives.

Given the challenges posed by aerial imagery—including variations in crosswalk size, shape, and environmental factors such as lighting and occlusions—specific modifications were introduced to enhance feature extraction through the integration of dual-branch pooling and advanced attention mechanisms. The backbone network serves as the primary feature extractor, processing input images to produce multi-scale feature maps. The neck network acts as an intermediary, refining and integrating features to enhance the detection of crosswalks across various scales and orientations. The detection head then utilizes these processed features to generate accurate predictions, culminating in the final model outputs.

To enhance the feature extraction capabilities of CrosswalkNet, two critical modules were incorporated: the Dual-branch Spatial Pyramid Pooling-Fast (SPPF) and the Soft-Convolutional Block Attention Module (Soft-CBAM). These enhancements aim to address the challenges posed by variations in crosswalk size, orientation, and environmental factors such as lighting and occlusions in aerial imagery.

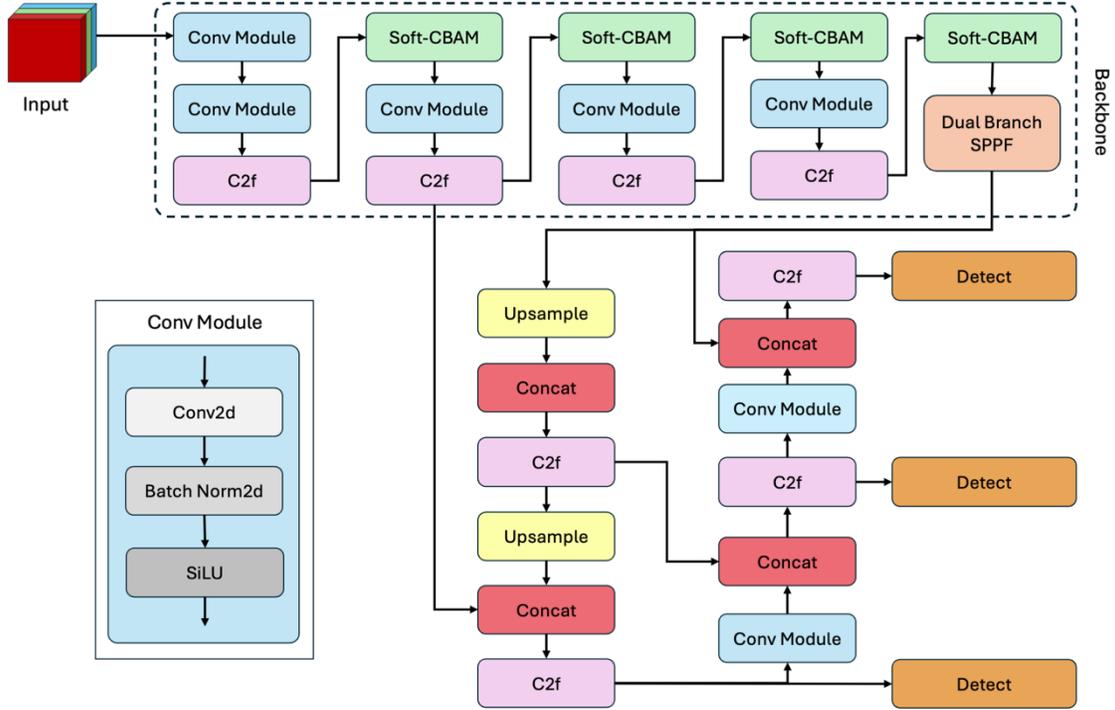

**Fig.2.** Architecture of CrosswalkNet.

The standard SPPF module, originally designed to aggregate multi-scale features and encode them into a fixed-length output regardless of input size, plays a crucial role in maintaining spatial hierarchies and improving the robustness of feature representation. However, the conventional SPPF module in YOLOv8 relies solely on max-pooling operations, which, although effective for achieving spatial invariance, often result in the loss of essential local information. To mitigate this issue, Wang et al., (2024) introduced an enhanced dual-branch SPPF module, which integrates soft-pooling alongside traditional max-pooling. This dual-branch approach, shown in **Fig.3**, optimizes feature utilization by selectively preserving pixels based on their weights in the feature map (Stergiou et al., 2021). As a result, improved scale invariance and richer feature representation are achieved, leading to enhanced detection performance. The mathematical formulation of the dual-branch SPPF is as follows:

$$Br_1 = Conv\left(Concat(Conv(x), MaxPool_1(x), MaxPool_2(x), MaxPool_3(x))\right) \quad (1)$$

$$Br_2 = Conv\left(Concat(Conv(x), SoftPool_1(x), SoftPool_2(x), SoftPool_3(x))\right) \quad (2)$$

$$Y = Concat(Br_1, Br_2) \quad (3)$$
8

Where, $MaxPool_i(x)$ refers to max-pooling at different scales, $SoftPool_i(x)$ represents soft-pooling operations to retain contextual information, and $Y$ denotes the concatenated feature map passed to the next stage.

The dual-branch SPPF module minimizes feature loss during the pooling process and improves the model's ability to analyze contextual data from diverse regions within an image, which is critical for accurately detecting crosswalks in complex environments.

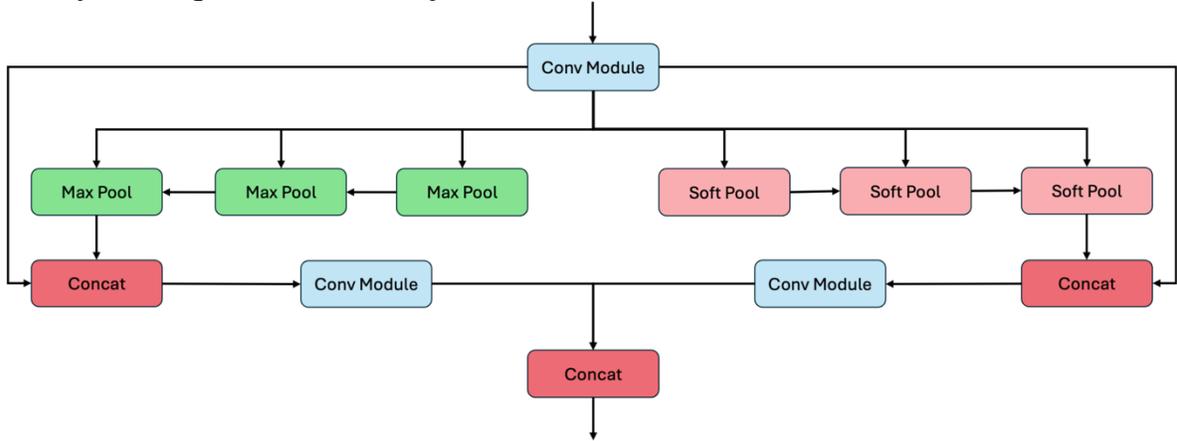

**Fig.3.** Dual-branch Spatial Pyramid Pooling – Fast.

Another key enhancement integrated into CrosswalkNet is the Soft-CBAM module (Wang et al., 2022), which enhances object detection performance by dynamically weighting feature representations, thereby improving object localization and identification. Unlike traditional Convolutional Block Attention Module (CBAM) approaches (Woo et al., 2018), which utilize average pooling, the proposed Soft-CBAM introduces soft-pooling, which selectively preserves more relevant features while reducing noise. As shown in **Fig.4**. Soft-CBAM applies both channel attention (CAM) and spatial attention (SAM) sequentially to focus on the most informative areas of the input feature map. The attention mechanism is mathematically expressed as follows:

$F_{att} = \sigma(W_s * F_{input} + W_c * F_{input})$  (4)

Where, $F_{input}$ is the input feature map, $W_s$ and $W_c$ represent spatial and channel attention weights, and $\sigma$ denotes the sigmoid activation function to scale attention scores.

The incorporation of Soft-CBAM enhances CrosswalkNet's ability to detect crosswalks under challenging conditions such as occlusions and varying lighting, ensuring more precise and reliable results.

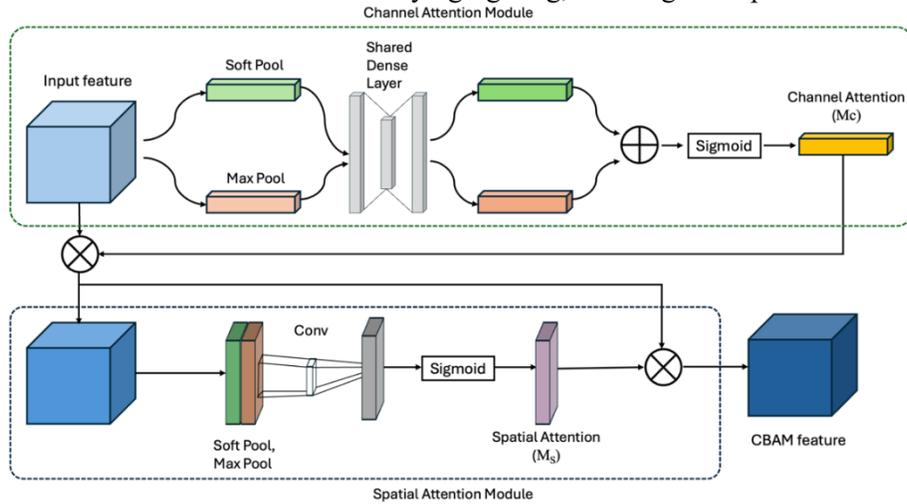

**Fig.4.** Soft-CBAM module.

## 5. Results and discussion

The performance of CrosswalkNet was systematically evaluated using multiple metrics, including precision, recall, and mean average precision (mAP), to assess its effectiveness in detecting pedestrian crosswalks from aerial imagery. The model's robustness was tested under diverse conditions such as occlusions, faded markings, and complex environmental settings, ensuring its applicability in real-world scenarios.

### 5.1. Model performance

Evaluating the detection performance of CrosswalkNet requires a comprehensive analysis of key metrics. Accuracy (Acc), precision (Pr), and recall (Re) were computed to quantify the model's ability to correctly identify crosswalks while minimizing false detections. The definitions of these metrics are as follows:

$$Acc = \frac{TP + TN}{TP + TN + FP + FN} \tag{5}$$

$$Pr = \frac{TP}{TP + FP} \tag{6}$$

$$Re = \frac{TP}{TP + FN} \tag{7}$$

where $TP, TN, FP$ and $FN$ represent true positives, true negatives, false positives, and false negatives, respectively. These metrics provide insight into the model's precision in distinguishing crosswalks from non-crosswalk areas and its effectiveness in capturing all relevant instances.

To further evaluate detection performance, mean average precision (mAP) was used, considering different Intersection over Union (IoU) thresholds. The mAP50 metric evaluates detection performance at a 50% IoU threshold, while mAP50-95 provides a more stringent evaluation by averaging precision across IoU thresholds from 50% to 95% in 5% increments, offering a thorough assessment of localization accuracy.

To ensure robust training and generalization, the dataset was divided into 82% for training and 18% for validation. The model was trained for 80 epochs, with evaluation performed periodically using the validation set to monitor accuracy, precision, and recall.

A cosine annealing learning rate scheduler was employed to enhance generalization and prevent overfitting. The learning rate cyclically decreased and increased, improving the model's ability to adapt to unseen data. Additionally, data augmentation techniques such as random rotations (±8 degrees), mosaic augmentation, and HSV (hue, saturation, and value) adjustments were applied to increase the model's robustness under varied conditions.

**Table 1** presents the results of baseline-Yolov8 models across different model configurations and input data types. It was observed that models trained on color (RGB) images consistently outperformed those trained on grayscale images, emphasizing the significance of color information for improving detection accuracy. Among all tested configurations, the YOLOv8-Large model demonstrated the best trade-off between precision, recall, and computational efficiency.

**Table 1**
Model performance considering all proposed enhancements.

| Dataset | Model | Precision | Recall | mAP50 | mAP50-95 |
|---|---|---|---|---|---|
| Grayscale | Nano | 0.938 | 0.846 | 0.947 | 0.719 |
|  | Small | 0.945 | 0.932 | 0.969 | 0.840 |
|  | Medium | 0.946 | 0.933 | 0.970 | 0.843 |
|  | Large | 0.946 | 0.932 | 0.970 | 0.844 |
|  | Extra-Large | 0.948 | 0.931 | 0.969 | 0.842 |
| Color (RGB) | Nano | 0.962 | 0.931 | 0.961 | 0.839 |
|  | Small | 0.957 | 0.931 | 0.965 | 0.858 |
|  | Medium | 0.965 | **0.933** | **0.969** | 0.867 |
|  | Large | 0.965 | **0.933** | **0.969** | 0.868 |
|  | Extra-Large | **0.972** | 0.932 | 0.966 | **0.869** |

Comparing the grayscale and RGB results presented in **Table 1** indicates that the inclusion of color information significantly enhances model performance across different model sizes. The YOLO models trained on the color dataset consistently achieved higher precision and mAP50-95 scores compared to their grayscale counterparts, underscoring the importance of color features in crosswalk detection. Notably, the YOLOv8 Extra-Large model demonstrated superior performance in terms of precision and mAP50-95, outperforming the Medium and Large models. However, it exhibited a slight decline in recall and mAP50, suggesting a trade-off between detection sensitivity and localization accuracy.

To achieve an optimal balance between detection performance and inference speed, the YOLOv8 Large model was selected as the baseline architecture for CrosswalkNet. This decision was based on its ability to provide high detection accuracy while maintaining computational efficiency, making it suitable for large-scale deployment. The results of the ablation study, as shown in **Table 2**, demonstrate the impact of the proposed architectural enhancements on model performance.

**Table 2** reveals that the incorporation of the Soft-CBAM module, Dual-branch SPPF, and cosine annealing significantly improved the precision and recall of the baseline model. The final CrosswalkNet model achieved a precision of 96.5% and a recall of 93.3%, with an mAP50 score of 96.9%. These improvements highlight the effectiveness of the proposed enhancements in refining feature extraction and optimizing training dynamics.

Compared to the base YOLOv8 Large model and prior studies (Verma and Ukkusuri, 2024; Luttrell et al., 2024), the observed performance gains are substantial within the domain of deep learning-based object detection. It is important to emphasize that the base model was meticulously configured and trained, making such significant improvements uncommon in typical ablation studies. The results validate the effectiveness of the proposed modifications in enhancing detection accuracy while ensuring scalability for real-world applications.

**Table 2**
Ablation study results of CrosswalkNet enhancements using the YOLOv8 large model with color dataset

|  | Precision | Recall | mAP50 | mAP50-95 |
|---|---|---|---|---|
| (a) = Base YOLOv8 Large model | 0.954 | 0.914 | 0.957 | 0.846 |
| (b) = (a) + Soft-CBAM | 0.959 | 0.920 | 0.951 | 0.849 |
| (c) = (b) + Dual-branch SPPF | 0.961 | 0.927 | 0.962 | 0.855 |
| (d) = (c) + cosine annealing | **0.965** | **0.933** | **0.969** | **0.868** |

**Fig. 5** illustrates examples of OBB detections, featuring both ladder and parallel types of crosswalks. To enhance clarity and detail, the images are selectively cropped to closely focus on the crosswalk areas. Notably, the model effectively detects crosswalks located in shadowed areas, such as those cast by buildings, and accurately identifies crosswalks adjacent to no parking hash markings. The model demonstrates versatility by accurately detecting crosswalks of varying orientations and lengths across diverse environmental settings, further illustrating its robustness and adaptability to complex urban landscapes.

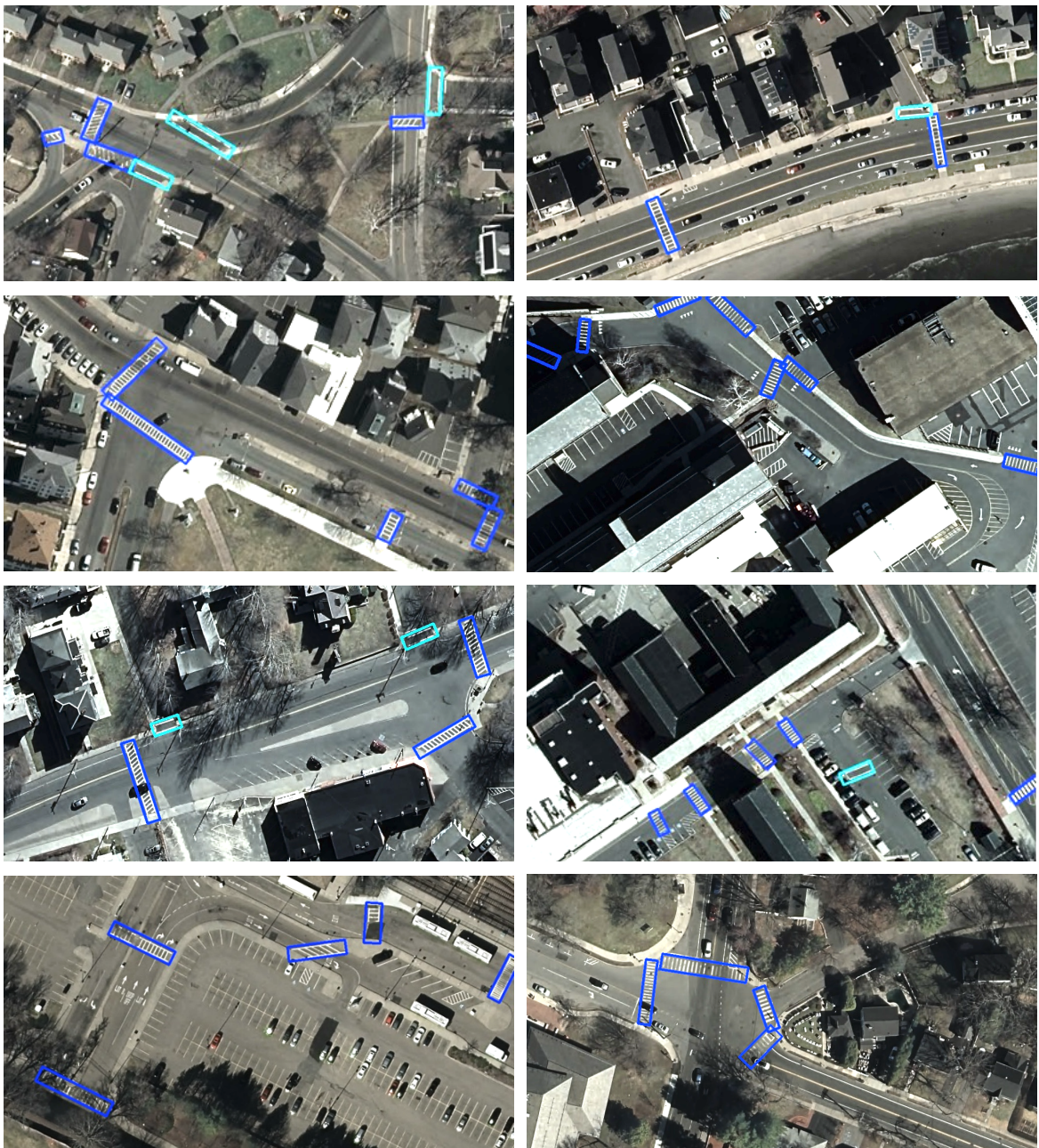

**Fig. 5.** Sample outputs of crosswalk in different orientations.

## 5.2. Sample results and discussion

The enhanced CrosswalkNet model, based on the YOLOv8-Large architecture, was utilized to detect pedestrian crosswalks from RGB aerial images of Massachusetts. Several challenges, such as shadows cast by buildings, partial occlusions from trees, and faded crosswalk markings, were encountered, which could obscure or alter their appearance. As illustrated in **Fig.6**, crosswalks were effectively detected under these challenging conditions. This performance can be attributed to the extensive and diverse dataset used for training, which included various crosswalk appearances under different conditions such as occlusions, wear, and environmental variations. As a result, the model was able to adapt effectively to fluctuations in visibility and environmental factors.

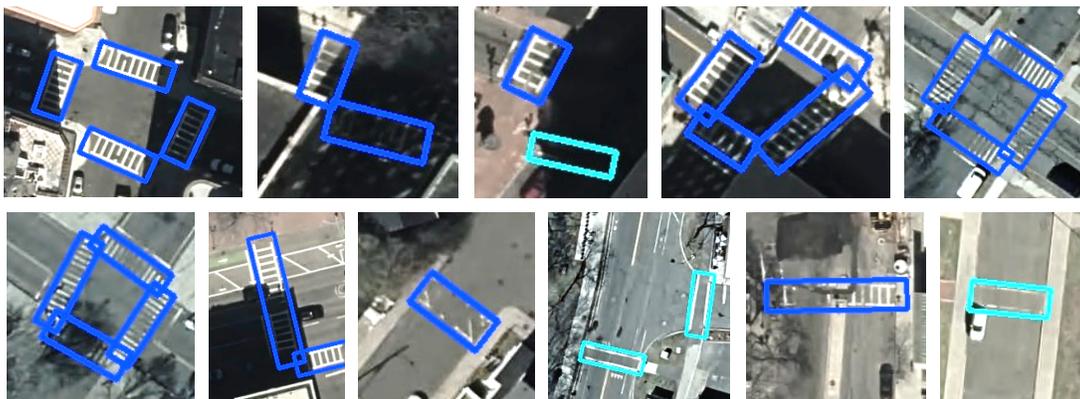

**Fig.6.** Examples of crosswalks detected in shaded, occluded conditions, or with faded markings.

Parking lots were identified as an additional challenge due to the presence of various markings, such as no-parking hash lines, lane dividers, and directional arrows, some of which visually resemble pedestrian crosswalks. These similarities can lead to misclassifications; however, as shown in **Fig. 7**, CrosswalkNet was able to successfully differentiate crosswalks from other pavement markings, ensuring high detection accuracy even in complex environments.

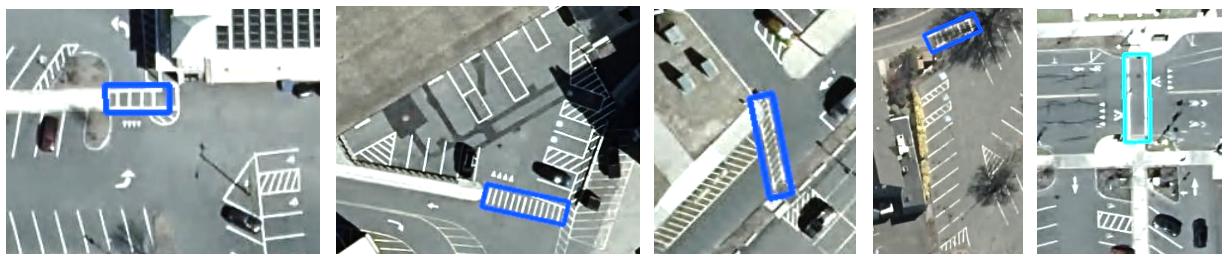

**Fig. 7.** Sample detections in parking lots.

To further evaluate the generalization capability of CrosswalkNet, aerial images with a resolution of 15 cm from Virginia, New Hampshire, and Maine were analyzed. Although the model was trained exclusively on data from Massachusetts, crosswalks in these new geographic regions were accurately detected without the need for additional fine-tuning. The successful application of the model across different geographic areas demonstrates its robustness and scalability, suggesting its suitability for deployment in diverse urban and suburban environments. Sample detection results from Maine, New Hampshire, and Virginia are presented in **Fig. 8**, further confirming the model's adaptability and consistent performance.

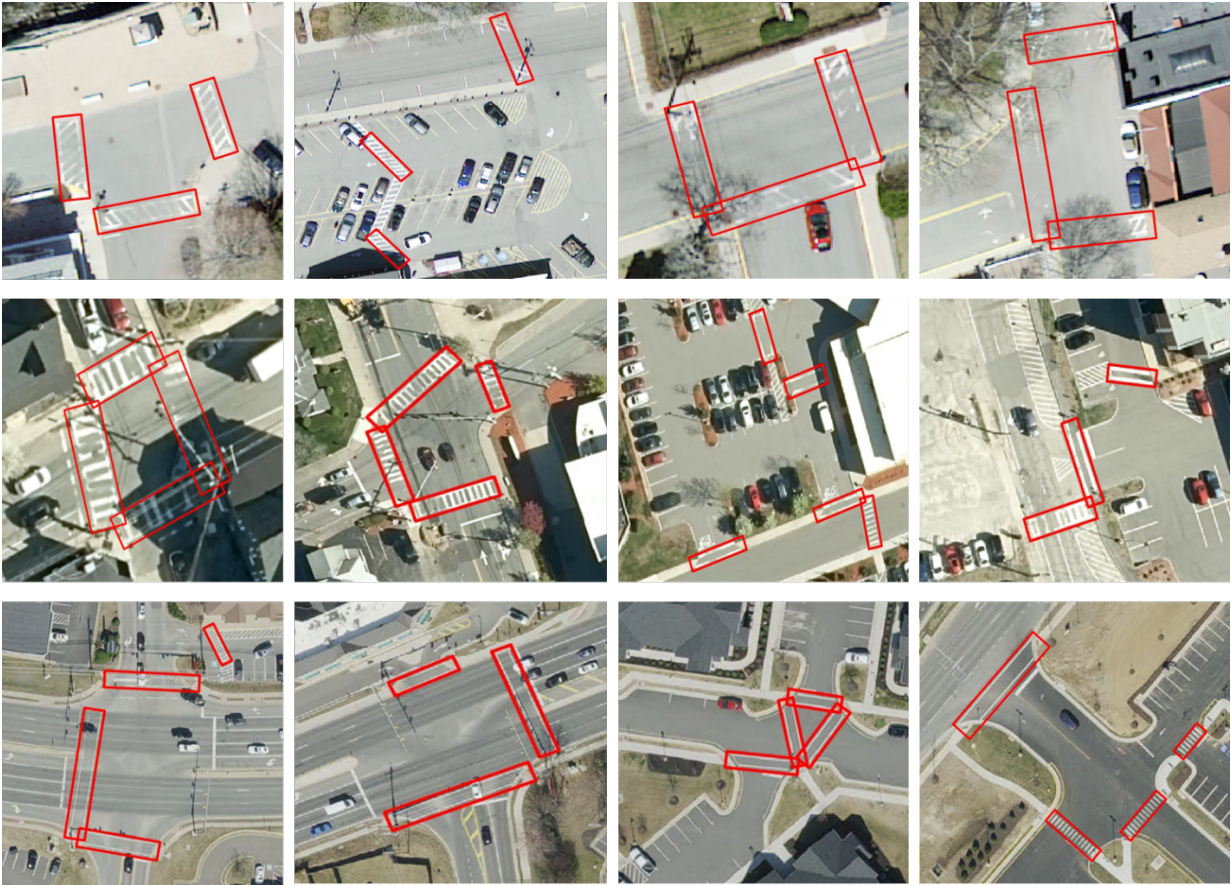

**Fig. 8.** Examples of detections on image data from other states (Top: Bar Harbor ME; Middle: Nashua, NH; Bottom: Garrisonville, VA).

## 6. Conclusions

This study introduced CrosswalkNet, an advanced deep learning framework for automated pedestrian crosswalk detection in high-resolution aerial imagery. CrosswalkNet effectively addresses key challenges associated with crosswalk detection, including variations in orientation, lighting, and occlusions, through the integration of several optimization strategies tailored for large-scale datasets.

One of the key contributions of this work is the integration of oriented bounding boxes (OBBs), which enables CrosswalkNet to detect crosswalks at varying orientations, significantly enhancing detection accuracy compared to traditional horizontal bounding boxes. Additionally, architectural enhancements were implemented through the incorporation of the Soft-CBAM module and the dual-branch SPPF, which improved feature extraction and refined spatial attention, resulting in higher precision and recall. The deployment of CrosswalkNet on high-performance computing (HPC) platforms has further improved scalability and efficiency, allowing for the efficient processing of extensive aerial datasets and facilitating near real-time analysis for safety and planning applications. Furthermore, a comprehensive dataset comprising over 23,000 annotated crosswalk instances was developed to support robust model training and validation, ensuring strong generalization capabilities across diverse environments.

The best-performing CrosswalkNet model achieved a precision of 96.5% and a recall of 93.3%, with an mAP50-95 of 86.8%, demonstrating its accuracy and effectiveness. The model's strong generalization ability was validated by its successful application to aerial images from Virginia, New Hampshire, and Maine without additional fine-tuning, underscoring its robustness across diverse geographic areas.

Additionally, the detection results, provided in polygon shapefile format, can be seamlessly integrated into GIS-based safety analysis, transportation planning, and urban mobility applications. This research highlights the potential of deep learning in improving pedestrian safety and accessibility through automated and scalable detection frameworks. In future studies, efforts will focus on expanding the framework to detect other pedestrian infrastructure elements such as sidewalks and pedestrian signals. Moreover, an exploration of crosswalk condition assessment will be conducted to support infrastructure maintenance planning. The adoption of self-supervised learning approaches and domain adaptation techniques will also be explored to further enhance the model's adaptability to different geographic regions and imaging conditions.

**Acknowledgements**

This study was undertaken as part of the Massachusetts Department of Transportation Research Program with funding from the Federal Highway Administration State Planning and Research funds. The authors are solely responsible for the facts, the accuracy of the data and analysis, and the views presented herein.

Special thanks to Kylie Braunius and Bonnie Polin for initiating this idea.

Xing, L., He, J., Abdel-Aty, M., Cai, Q., Li, Y., Zheng, O., 2019. Examining traffic conflicts of up stream toll plaza area using vehicles' trajectory data. Accident Analysis and Prevention 125, 174–187. https://doi.org/10.1016/j.aap.2019.01.034

Yang, X., Li, X., Ye, Y., Lau, R.Y.K., Zhang, X., Huang, X., 2019. Road Detection and Centerline Extraction Via Deep Recurrent Convolutional Neural Network U-Net. IEEE Transactions on Geoscience and Remote Sensing 57, 7209–7220. https://doi.org/10.1109/TGRS.2019.2912301

Yi, J., Wu, P., Liu, B., Huang, Q., Qu, H., Metaxas, D., 2021. Oriented object detection in aerial images with box boundary-aware vectors. Proceedings - 2021 IEEE Winter Conference on Applications of Computer Vision, WACV 2021 2149–2158. https://doi.org/10.1109/WACV48630.2021.00220

Zhai, S., Shang, D., Wang, S., Dong, S., 2020. DF-SSD: An Improved SSD Object Detection Algorithm Based on DenseNet and Feature Fusion. IEEE Access 8, 24344–24357. https://doi.org/10.1109/ACCESS.2020.2971026

Zhai, Y., Cui, G., Gu, Q., Kong, L., 2015. Crosswalk Detection Based on MSER and ERANSAC. IEEE Conference on Intelligent Transportation Systems, Proceedings, ITSC 2015-Octob, 2770–2775. https://doi.org/10.1109/ITSC.2015.448

Zhang, Z. De, Tan, M.L., Lan, Z.C., Liu, H.C., Pei, L., Yu, W.X., 2022. CDNet: a real-time and robust crosswalk detection network on Jetson nano based on YOLOv5. Neural Computing and Applications 34, 10719–10730. https://doi.org/10.1007/s00521-022-07007-9
19